# Event-Driven Time Series Analysis and the Comparison of Public Reactions on COVID-19

Md. Khayrul Bashar

Tokyo Foundation for Policy Research, Tokyo 106-6234, Japan

## Abstract

*The rapid spread of COVID-19 has already affected human lives throughout the globe. Governments of different countries have taken various measures, but how they affected people lives is not clear. In this study, a rule-based and a machine-learning based models are applied to answer the above question using public tweets from Japan, USA, UK, and Australia. Two polarity timeseries (meanPol and pnRatio) and two events, namely "lockdown or emergency (LED)" and "the economic support package (ESP)", are considered in this study. Statistical testing on the sub-series around LED and ESP events showed their positive impacts to the people of (UK and Australia) and (USA and UK), respectively unlike Japanese people that showed opposite effects. Manual validation with the relevant tweets shows an agreement with the statistical results. A case study with Japanese tweets using supervised logistic regression classifies tweets into heath-worry, economy-worry and other classes with 83.11% accuracy. Predicted tweets around events re-confirm the statistical outcomes.*

## 1. Introduction

Now a days, the microblogging platforms especially the Twitter has become essential tools for communication, especially for political or professional leaders including health professionals and public to interact with each-other as well as their intra-domain communication [1]. It has become a popular platform starting from the third world countries to the developed countries. This platform is playing active roles to disseminate public health information and to obtain real-time health data using crowdsourcing methods. It has already been used to disseminate information during many public health disasters such as influenza in 2009, the outbreak of Ebola virus (EV) in 2014, the spread of Middle Eastern respiratory syndrome in 2015 and the outbreak of Zika virus in the late 2015 [2]. World Health Organization (WHO) and the Centers for Disease Control and Prevention (CDC) have adopted the use of twitter and other social media by realizing the twitter's potential to inform and educate the public and governmental agencies. Several systematic review papers identified six main uses of Twitter for public health: (i) analysis of shared content, (ii) surveillance of public health topics or diseases, (iii) public engagement, (iv) recruitment of research participants, (v) public health interventions, and (vi) the network analysis of Twitter users [3]. On the other hand, the Twitter platform has been facilitating the analysis of many political issues such the prediction of vote percentage, the political campaign and its effects, the analysis of political homophily, the detection of election fraught etc. [4]. Some researchers are using corona related tweets for political framing [5].

In December 2019, the first diagnosis of a novel coronavirus, formally named severe acute respiratory syndrome coronavirus 2 (SARS-CoV-2), was made in the Wuhan City, Hubei Province, China. Later, it's rapid spread has drawn increasing media and public attention. Press coverages have further elevated in January 21, 2020 when the CDC activated its emergency operations center and the WHO began publishing daily situation reports. Subsequent travel bans, large-scale quarantine of Chinese residents generated significant interests by the public. Local and global media have gradually become more active to update the ongoing corona situations and to publish in-depth analysis on corona pandemic. Governments of different countries have gradually taken various measures including the declaration of emergency or lockdown and the financial support packages for flattening the peak situations of corona virus. At the same time, the public health researchers have taken many emergency projects to find the source of corona virus and to discover vaccines for controlling its spread [6]. However, there is limited insight on how the public sentiments are affected by the corona severity and/or the various government actions such as the declaration of lockdown or emergency conditions and the economic support packages.

## 2. RELATED WORKS

Over the past months after the onset of coronavirus, several works were published. Three main streams of works are progressing: (i) the development of models for estimating the spread of corona virus and the associated infection or death cases (ii) the development of vaccine as a remedy against this deadly disease, and (iii) the analysis of epidemic's impacts on the public health, economy and the global supply systems. Several authors have proposed advanced predictive models based on genetic programming and advanced machine learning algorithms including deep learning (e.g., LSTM) to address issues in the first stream [7] [8]. These models help to interpret patterns of the public sentiments in disseminating the related health information and assess political and economic influences of the spread of the virus. Researchers in the biological and medical domains are handling the second stream and are actively working on the vaccine development [6]. At present, several vaccine candidates are in the market places and several others are in the trial stage. Several researches were published in the third stream in which lexicon-based approaches, machine learning, and topic models are used [3] [9] [10]. These studies revealed the economic and political impacts of the COVID-19 as the most commonly discussed topics, while the risk for the public health and its prevention were the least discussed topics. Another aspect in this stream is the analysis of retweet networks and retweet speed. Beside the scientific researches, the governments in different countries have been taking various preventive measures such as the declaration of lockdown or emergency conditions, the financial support packages for the individual and business supports etc. However, it is quite unclear whether these measures really affected the public lives and sentiments. Moreover, how the public worries towards the heath and economy has changed or is changing over time is also not fully explored. In this study, we therefore focus on exploring these issues. For the first one, we will consider two events namely the first declaration of lockdown or emergency (LED) and the economic support package (ESP). In the second case, a supervised machine learning method will be developed to classify public reactions into health and economy worries and their progression over time. Such understanding would enable the largescale opportunities for the prediction of public worries towards the health and economy during future catastrophic events.

## 3. DATA COLLECTION

### 3.1. Dataset-1

To test our research questions on the corona pandemic, we collected public tweets of the four countries (Japan, USA, UK, and Australia) for a six-month duration (January 2020 to June 2020). Target countries were selected considering geographical locations and the nature of infection varieties due to corona virus. Tweets were collected using "keyword" based search strategy [1] [11] [12]. Since we are interested in the public reactions in relation to the respective government actions, we used leader's twitter handle as one of the search terms. For example, to extract public tweets for Japan, we constructed the query-string by concatenating corona related terms with the twitter handle of the corona in-charge in Japan (i.e., Economy Minister Yasutoshi Nishimura; @nishy03) by using logical "AND" operation.

In the keyword-based searching, several keywords (or hashtag keywords) are used for downloading the required tweets. The selection of keywords and their numbers usually depends on the target of projects [3] [13] [14]. Our study aims at comparing public reactions to COVID-19 among three English-spoken and one non-English spoken (e.g., Japan) countries. In this study, we therefore selected seven commonly used keywords and/or hashtags in the selected countries: "corona", "coronavirus", "novel coronavirus", "COVID-19", "COVID19", "virus", and "covid". During the selection of the above keywords, we were motivated by the Oxford English Dictionary (OED) ranking and several other sources, namely Instagram, Yale Medicine Team, and several online reports [15 - 21]. With the mentioned keywords, we finally extracted 28, 930 public tweets using twitter standard search API for the mentioned four countries. A threshold of 100 maximum tweets per day was set to make the API workable during data collection. The characteristics of our datasets are given in given in Table 1. The number of the collected tweets is approximately proportional to the number of active twitter users (on the corona issue) in each country. Although, the dataset is not very large one, the random selection of the tweets by the API greatly reduces the possible biases on the collected datasets. These datasets will be used for statistical analysis of events. Besides, the dataset for the Japanese public will also be used for the classification of the public reactions in Japan.

Table 1: Information on public tweet datasets

| Group | Total days | Min | Max | Avg | Tweets |
|---|---|---|---|---|---|
| Japan | 145 | 1 | 100 | 28.08 | 4072 |
| USA | 173 | 1 | 100 | 88.09 | 15240 |
| UK | 163 | 1 | 100 | 63.09 | 8090 |
| Australia | 164 | 1 | 81 | 16.93 | 1528 |

### 3.2. Dataset-2

It is a small dataset, constructed for Japan based on a different set of keywords than those used for collecting Japanese tweets in Dataset-1. These keywords were selected to directly construct annotated datasets for supervised classification of two dominant classes, i.e., health-worry ("hWorry"), and economy-worry ("eWorry") as found in the Dataset-1. The translated version of some Japanese keywords for "hWorry" and "eWorry" classes include mask shortage, medical system collapse, lack of testing, bankruptcy, corona recession, damage on tourism etc. To keep consistent with the tweet categories in the Dataset-1, we have created a third class, designated as "other", by including freely available USA airline review tweets [22]. Therefore, this dataset consists of 146, 349, 300 samples for "hWorry", "eWorry", and "other" classes, respectively. Finally, the Dataset-2 is used for training and validation, while the Japanese tweets in Dataset-1 is used as the final testing sets in our study.

## 4. EVENT-DRIVEN SENTIMENT ANALYSIS

### 4.1. Overview

In this study, timeseries tweet data is first preprocessed to eliminate URL, punctuations, stop words and rare words. Lemmatization is also performed to remove inflectional endings only and to return the base or dictionary form of a words. Then sentiment parameters are extracted and the date-wise sentiment timeseries are constructed for mean polarity ("meanPol") and positive-negative count ratio ("pnRatio") using a rule-based sentiment extraction model, called VADER [23]. To explore whether government actions against corona pandemic has potential effects on the public sentiments, we consider two well-known events, namely the first declaration of lockdown or emergency conditions (LED), and the first declaration of economic support package (ESP) among four countries Japan (JAP), USA, UK, and Australia (AUS). Statistical error analysis and Welch's *t*-testing has been performed fifteen days before and after each event for each country to justify the significance of the actions taken by each government. A validation study on the event-related tweets has been performed to verify the results of the statistical analysis. Finally, a case study using the Japanese tweets has been performed which employs logistic regression to classify and then to verify the tweets related to "hWorry", "eWorry", and the "other" categories, respectively. Results showed the promising performance of the proposed method and analysis. Figure 1 below show the overview of our approach.

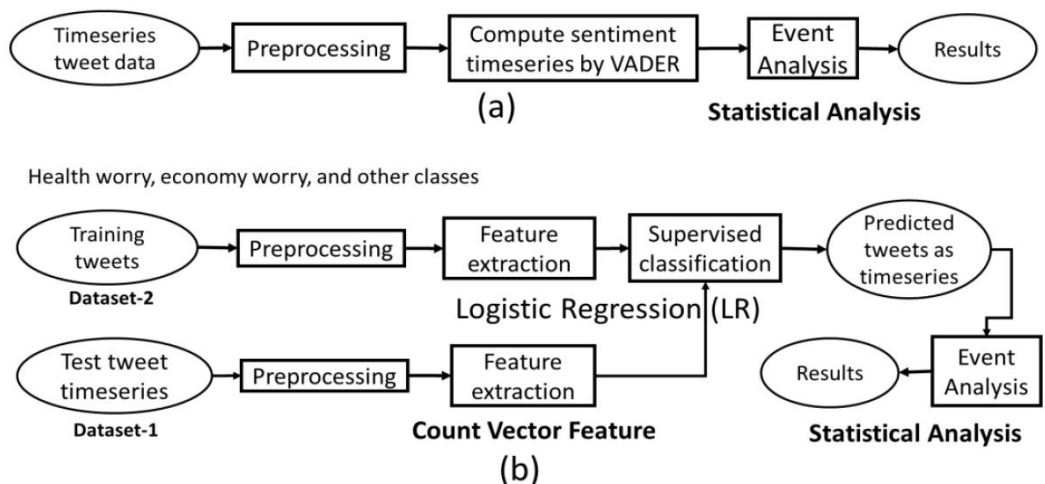

Fig. 1. (a)Even-driven Processing of twitter sentiment timeseries (b) Classification of public reactions to heath worry, economic worry, and others.

### 4.2. Polarity Extraction

A rule-based model, entitled **Valence Aware Dictionary and Sentiment Reasoner** (VADER)**,** is adopted for public sentiment analysis from the twitter data. This model combines lexical features with five general rules that embody grammatical and syntactical conventions for emphasizing sentiment intensity and showed effectiveness in social media data analysis. It tells not only about the positivity and negativity score but also tells us about how positive or negative a sentiment is. It outperforms individual human raters and captures better contexts compared to eleven benchmarks including ANEW, SentiWordNet, and machine learning oriented techniques relying on Naive Bayes, Maximum Entropy, and Support Vector Machine (SVM) algorithms. Please refer to [23] for detail information.

### 4.3. Sentiment Parameters

We have computed several sentiment parameters using VADER. These are mean polarity ("meanPol") and the positive-negative polarity count ratio ("pnRatio") in our analysis. In VADER, emotion intensity or sentiment score is measured on a scale from -4 to +4, where -4 is the most negative and +4 is the most positive. The midpoint 0 represents a neutral sentiment. VADER can map emoticons that appear in the social media texts like tweets. In our study, we only consider positive and negative scores of each sentiment bearing word. The above parameters were computed based on compound polarity score as defined by Eq. 1.

$$Compound\ Score = \frac{x}{\sqrt{x^2 + \alpha}} \qquad (1)$$

where x is the sum of the sentiment scores of the constituent words of the sentence or sentences and α is a normalization parameter that we set to a default value 15. This gives us a normalized score between -1 (most extreme negative) and +1 (most extreme positive). Please refer to (Hutto et al., 2014) for more information.

### 4.4. Classification of Public Reactions

Understanding the heath and economic related public reactions is very important in decision making by the leaders and government of a country. This could help proactive policy decisions especially during disastrous situation like corona pandemic. In this study, we have developed a classification algorithm using logistic regression to classify public reactions as "hWorry" and "eWorry" using Japanese tweets as a case study.

After preprocessing, the count-vectorizer are applied to construct numerical feature matrix for classification [24]. Classical machine learning models usually perform better than the advanced deep learning models with relatively small datasets having short-length text data. After an investigation with Dataset-2 using four machine learning models namely the Naïve Bayes (NB), linear support vector machine (LSVM), logistic regression (LR), and Random Forest (RF), we have finally selected the count-vector feature and the LR model for our analysis. The collection of Japanese tweets for six-month duration (Please refer to Section 3.1) was used as the final testing set. Japanese tweets are translated into English using "Googletrans", a python library that implemented Google Translate API ("Googletrans", a python library). Classification performance is computed using well-known evaluation metrics as defined below [25].

$$PREC = \frac{TP}{TP + FP} \qquad (2)$$

$$SN = \frac{TP}{TP + FN} \qquad (3)$$

$$SP = \frac{TN}{TN + FP} \qquad (4)$$

$$ACC = \frac{(TP + TN)}{(TP + TN + FP + FN)}, \qquad (5)$$

where TP, FN, FP and TN represent the number of true positives, false negatives, false positives and true negatives, respectively.

## 5. EXPERIMENTAL RESULTS AND DISCUSSION

### 5.1. Polarity Timeseries Analysis

With the collected data, we have generated "meanPol" and "pnRatio" timeseries for four countries (Figs. 2 and 3). Figure 2 shows that the Japan, UK, and Australia have large variations in the mean polarity before March 15, 2020, while the mean polarity for USA varies wildly throughout the six-month duration. The mean polarity for the Australia also varies widely after the 2$^{nd}$ week of April 2020. The most negative polarity was observed for UK and AUS public, while the most positive polarity was found for Japanese people. An approximately similar trend was observed in the pnRatio timeseries as shown in Fig. 3.

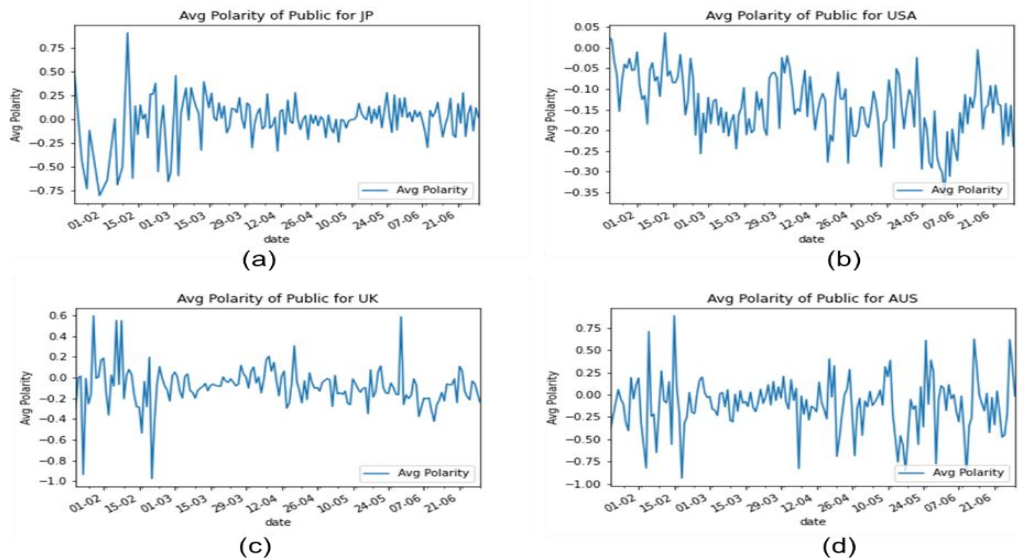

Fig.2. Plots for average polarity for (a) Japan, (b) USA, (c) UK, and (d) Australia

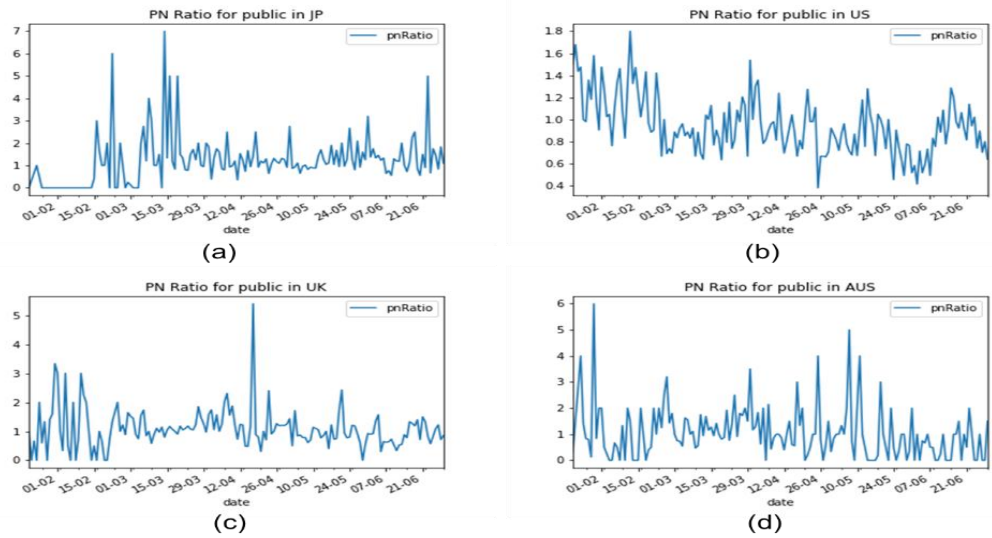

Fig.3. Plots for pnRatio for (a) Japan, (b) USA, (c) UK, and (d) Australia

## 5.2. Event Analysis

To address the research question, we have considered two events: (i) the first declaration of lockdown or emergency (LED) and (ii) the first declaration of economic support package (ESP).

**LED**: The starting date at which some restrictions on the human and/or business activities at public places were first imposed for the sake of ensuring public and national safety in a special occasion (e.g., violence, pandemic etc.). In case of the corona pandemic, such event is known as lockdown or the state of nation-wide emergency declaration. In this study, we considered this date as an important event.

**ESP**: On the eve of disaster situations, the national government usually declares financial support to protect its citizens from financial and mental crisis. During COVID-19, most of the affected countries declared such packages at different point of time. In our study, the first occurrence of this kind of support is considered as an event for analysis.

Table 2. MeanPol and pnRatio before and after LED

|  | **MeanPol** |  | **pnRatio** |  |
| --- | --- | --- | --- | --- |
|  | BeforeLED | AfterLED | BeforeLED | AfterLED |
| Japan | 0.039362 | 0.003026 | 1.430606 | 1.282577 |
| USA | -0.16379 | -0.14241 | 0.876282 | 0.925935 |
| UK | -0.06723 | 0.013028 | 1.045158 | 1.429418 |
| Australia | -0.10858 | -0.01557 | 1.073405 | 1.536419 |

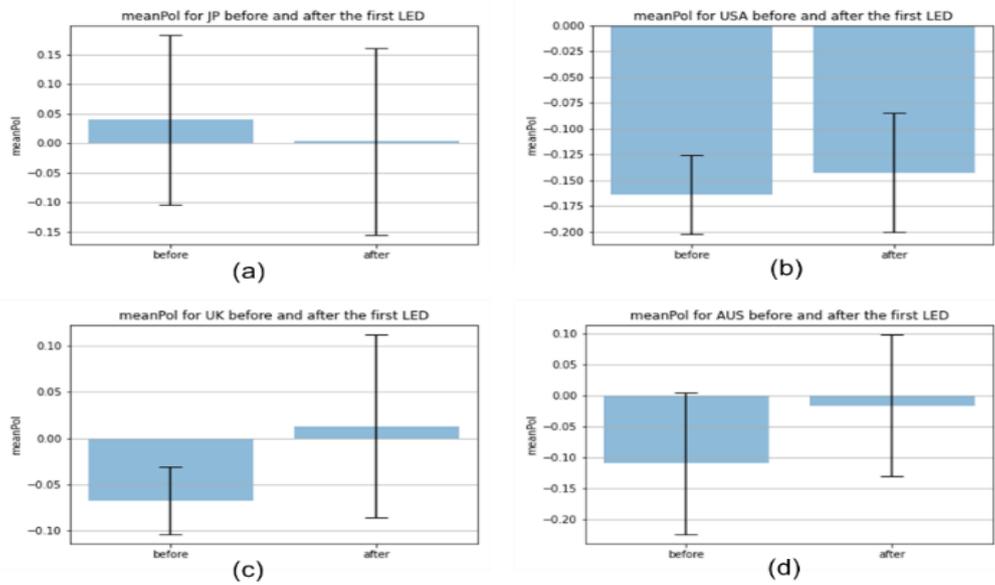

Fig.4. Error plots of meanPol for LED. (a) Japan, (b) USA, (c) UK, and (d) Australia

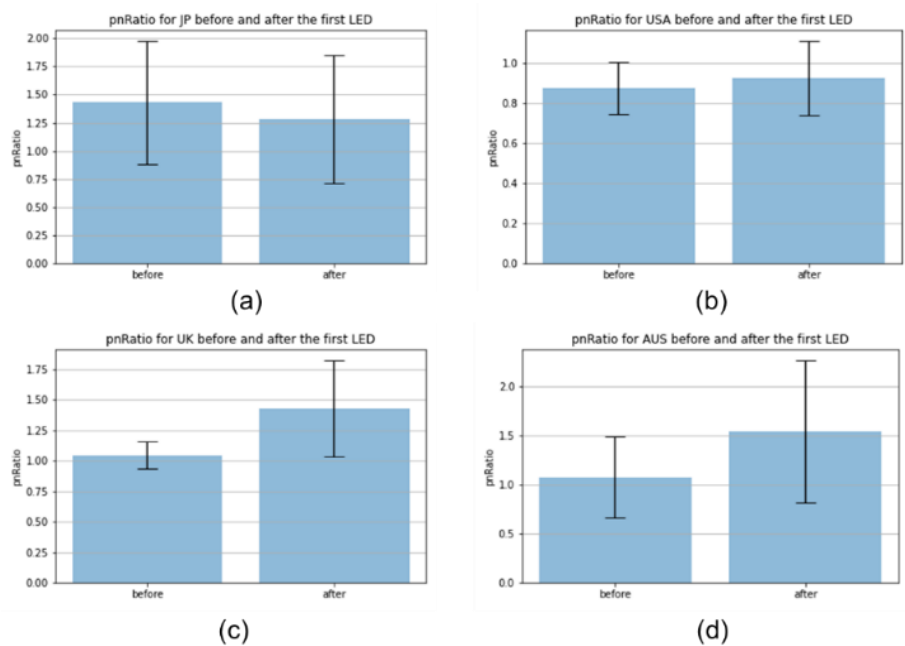

Fig.5. Error plots of *pnRatio* for LED. (a) Japan, (b) USA, (c) UK, and (d) Australia

Table 2 and Figs. 4 and 5 show the meanPol and pnRatio parameters 15 days before and after the LED event. These figures clearly show that except for Japan, the negative meanPol reduces after the LED event for Australia, UK, and USA. On the other hand, pnRatio increases after the LED event for the same three countries (Australia, UK, USA) except Japan. This indicates that the public in those three countries somewhat accepted the LED as a positive step. While in Japan, it showed the opposite behavior indicating that the Japanese people are not much satisfied with the LED implementation.

Table 3 and Figs. 6 and 7 show the meanPol and pnRatio parameters 15 days before and after the ESP event. Figures show that except for Japan and Australia, the negative meanPol reduces after the ESP event for UK, and USA. On the other hand, pnRatio increases after the ESP event for the same two countries (UK, USA) except the Japan and Australia, which showed the opposite behavior. This indicates that the public in the USA and UK accepted the ESP as a positive step to some extent. While the public in Japan and Australia are not much satisfied with the ESP implementation.

Table 3. MeanPol and pnRatio before and after ESP events

|  | **MeanPol** |  | **pnRatio** |  |
| --- | --- | --- | --- | --- |
|  | BeforeESP | AfterESP | BeforeESP | AfterESP |
| Japan | 0.039362 | 0.003026 | 1.430606 | 1.282577 |
| USA | -0.16379 | -0.09534 | 0.876282 | 1.044415 |
| UK | -0.06723 | -0.00656 | 1.045158 | 1.292268 |
| Australia | -0.10858 | -0.04912 | 1.073405 | 1.292818 |

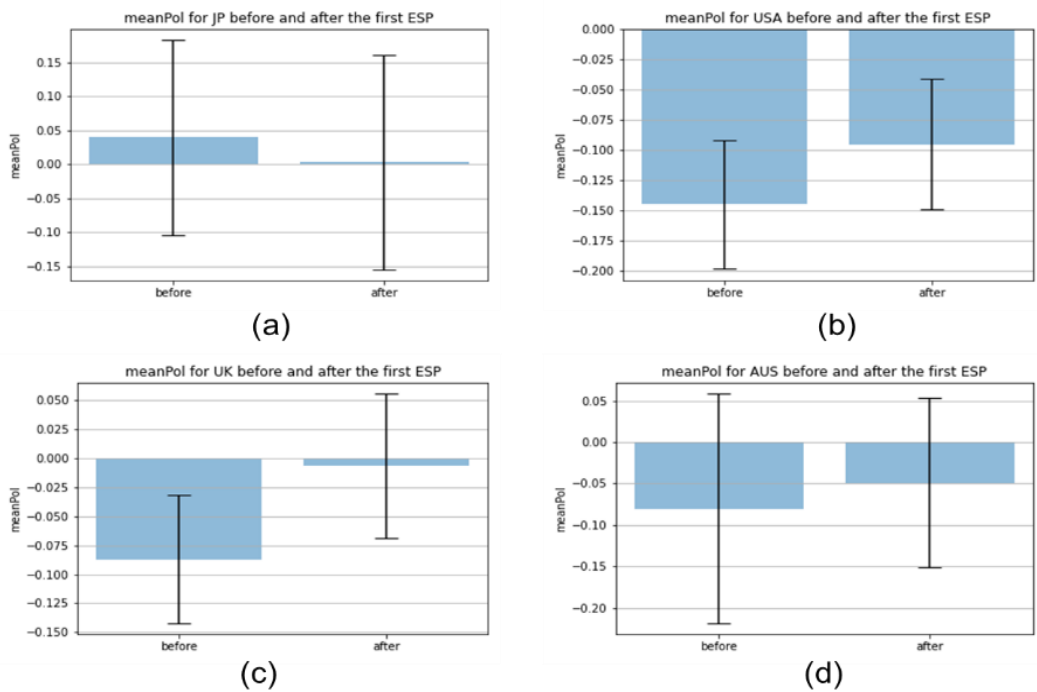

Fig.6. Error plots of *MeanPol* for ESP. (a) Japan, (b) USA, (c) UK, and (d) Australia

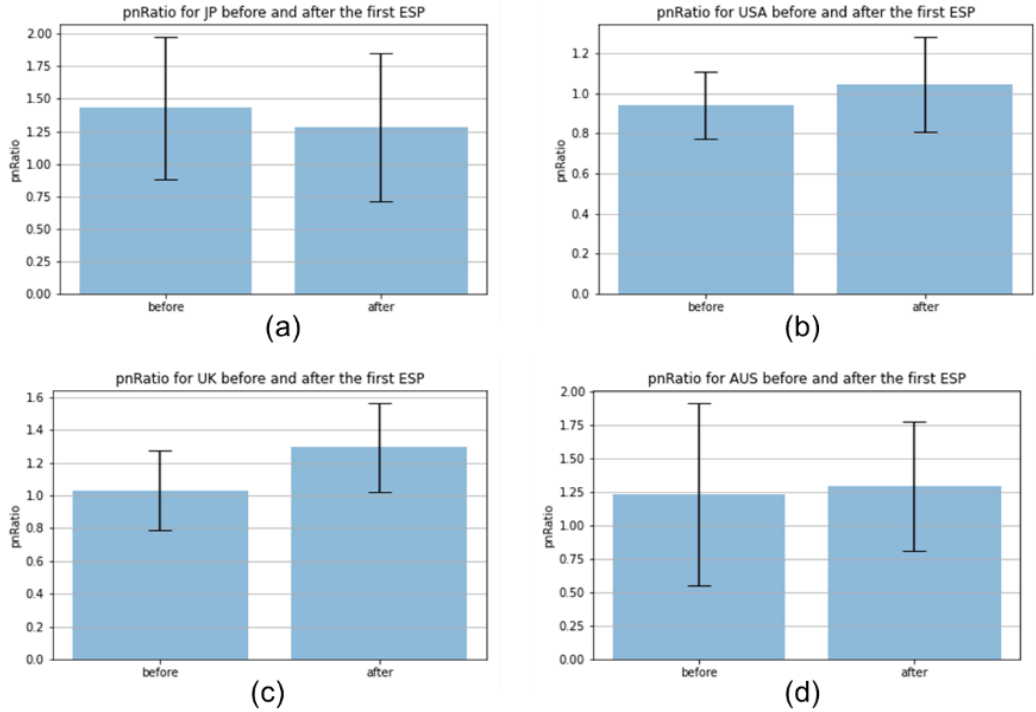

Fig.7. Error plots of *pnRatio* for ESP. (a) Japan, (b) USA, (c) UK, and (d) Australia

To justify the above observations more closely, we have performed Welch's *t*-testing on the 15-days sentiments before and after the mentioned two events. Table 4 shows the test results. Results showed significant differences of the MeanPol and pnRatio parameters over the LED event for UK and Australia, while they showed significant differences for USA and UK on the FSP event. These testing outcomes are quite congruent with the results from the statistical error analysis (Figs. 6 and 7) except for the USA on the LED event. However, the Japanese tweets does not show any significant differences for both events. In conclusion, we can say that UK and USA people are somewhat satisfied with the government's ESP event, while the UK and Australian people are satisfied with the LED event.

Table 4. Welch's *t*-test Results for Event Sentiments Analysis

| Country | Parameters | LED | | ESP | |
|---|---|---|---|---|---|
| | | *MeanPol* | *pnRatio* | *MeanPol* | *pnRatio* |
| Japan | t | 0.6399 | 0.7023 | 0.6399 | 0.7023 |
| | p | 0.5274 | 0.4882 | 0.5274 | 0.4882 |
| | dof | 27.744 | 27.948 | 27.744 | 27.948 |
| USA | t | -1.155 | -0.8212 | -3.860 | -2.334 |
| | p | 0.2590 | 0.4192 | **0.0007** | **0.0291** |
| | dof | 24.262 | 25.188 | 25.116 | 21.845 |
| UK | t | -2.845 | -3.503 | -3.151 | -3.160 |
| | p | **0.0108** | **0.0029** | **0.0045** | **0.0052** |
| | dof | 17.748 | 16.173 | 22.669 | 18.529 |
| Australia | t | -2.152 | -2.074 | -1.448 | -1.292 |
| | p | **0.0401** | **0.0497** | 0.1587 | 0.2070 |
| | dof | 27.999 | 22.296 | 27.656 | 27.416 |

To justify the above observations, we have manually investigated the top 50 most negative tweets before and after each event. Negative tweets are selected based on the observations that the overall MeanPol is mostly negative for all countries except Japan which is slightly positive (Please refer to the Tables 2 and 3). Figures 8 and 9 show the results of investigation before and after the LED and ESP event, respectively.

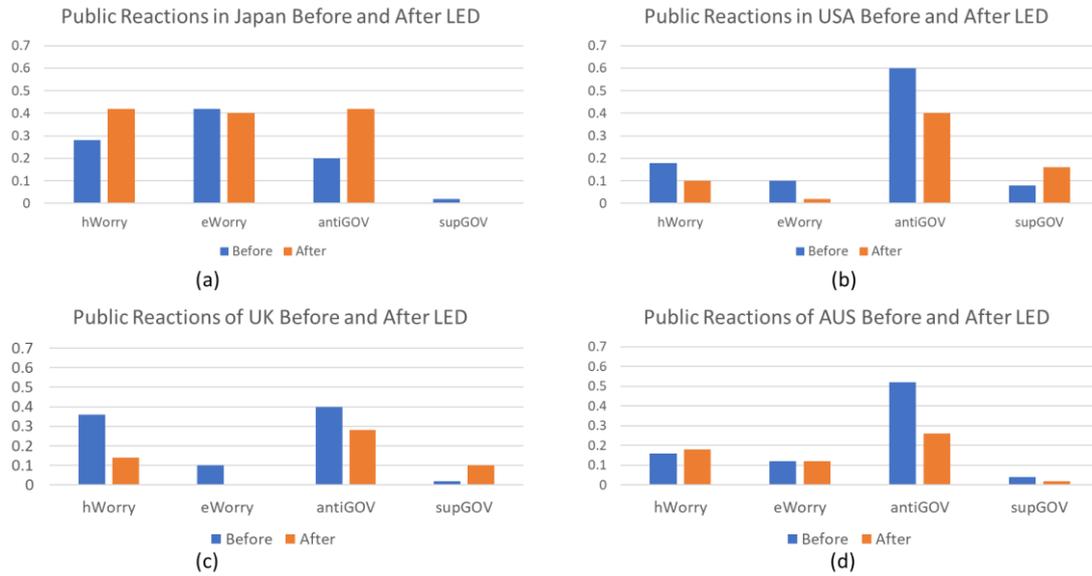

Fig.8. Number of tweets for LED. (a) Japan, (b) USA, (c) UK, and (d) Australia

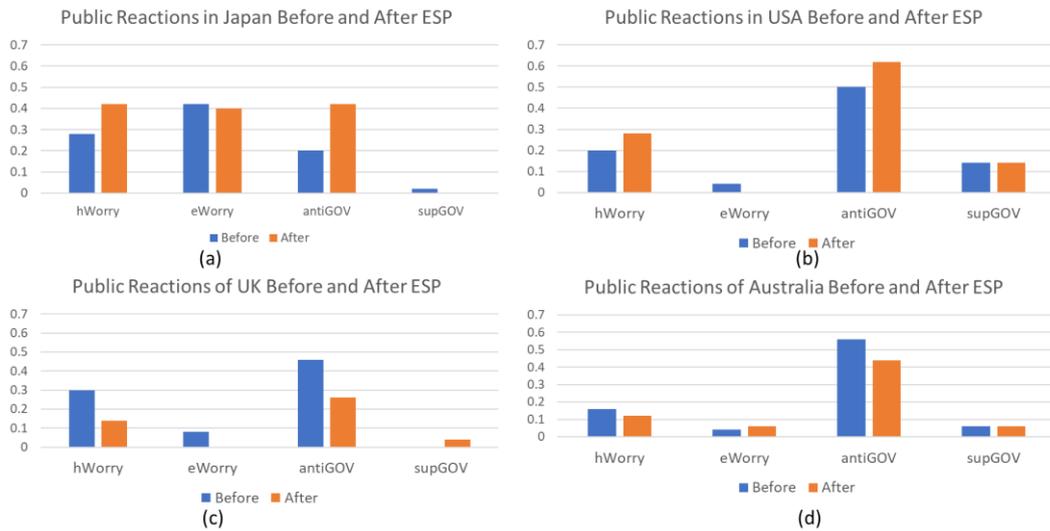

Fig.9. Number of tweets for ESP. (a) Japan, (b) USA, (c) UK, and (d) Australia

The depicted results in the above figures agree with the results of statistical analysis as explained above. The Fig. 8 (b-d) for the Australia, UK, and USA show similar tendency with decreasing worries for health and economy in case of the LED event, while Japan shows the opposite behavior. Note that Australia has a small perturbation for the health worry case. This is perhaps a side effect for not including all negative tweets in manual verification. Similar findings were observed in case of the ESP event (Fig. 9). In this case, the USA is not in agreement with the results of statistical analysis. Again, this perturbation may happen because of the mentioned reason as above as well as the limitation due to human annotation. However, the overall outcomes are approximately in agreement with the results that we obtain from statistical analysis. In addition, we can go more insight with the validation results in terms of anti-government public sentiments. The anti-government sentiments of the Japanese people have increased after each event, while the same for the most other countries have decreased except the USA with the ESP event. Results also showed the highest anti-government sentiments in USA and Australia compared to other countries. Note that both the LED and ESP events were declared in the same date in Japan. Note also that the Japanese people gave stronger reactions on the health and economic issues compared to other countries (Fig. 8 and 9 (a)).

### 5.3. Classification of Public Reactions

Based on the above observations, we have decided to perform a classification study on the reactions of Japanese public as a case study. In this case, supervised classification method has been developed as explained in the section 4.4. **Table 5** below shows the results of classification using count vectorizer with LR classification model. With the three classes ("hWorry", "eWorry", and "other"), we have obtained 83.11% average classification accuracy (macro average) with the high percentage of specificity (87.33%) and reasonable precision (74.78%) and sensitivity (74.66%). Note that the classification accuracy for the "hWorry" class is relatively low compared to the "eWorry" class, indicating the effects of our unbalanced training set. However, we hope to improve our results by adopting proportional weighting scheme or advanced deep learning model with larger dataset and more appropriate numerical feature model. Results also indicates that we can successfully classify high-level concepts like the worries for heath and economy using the properly designed training sets.

Table 5: Classification Performance on the test set using LR with count-vector feature

|  | PREC | SN | SP | ACC |
| --- | --- | --- | --- | --- |
| hWorry | 0.767 | 0.690 | 0.895 | 0.827 |
| eWorry | 0.735 | 0.860 | 0.845 | 0.850 |
| other | 0.742 | 0.690 | 0.880 | 0.817 |
| Macro average | 0.7478 | 0.7466 | 0.8733 | 0.8311 |

*Test set: 4072 tweets, verified 100 tweets from each class

Figures 10 shows per day classified tweets as the line plot. These plots showed clear dominance of economy worry of the Japanese people compared to their health worry especially between the beginning of April 2020 to the first week of May 2020. This was, in fact, the peak devastation period at which the Japanese government declared the emergency condition and the financial support package for the first time. Since the test-set has more than 4000 tweets, it is very time consuming to label all tweets manually. We therefore manually annotated the first 300 tweets (100 tweets per class) using the resultant tweet-groups after classification.

To justify the classification results against the "LED" and "ESP" events, the predicted tweets 15 days before and after each event were inspected manually. Results were shown in the Table 6 which shows that the counting and percentage of the "hWorry" tweets has increased after the LED and ESP events, while the "eWorry" tweets has slightly decreased after the declaration of each event. These results are consistent with the statistical analysis, shown in Figs. 8 (a) and 9(a). This observation once again validates that the Japanese people were not very happy with the government corona measures. Many inspected tweets also revealed that the people were unhappy with the delay in implementing the necessary health and economic measures of the government as well as its bureaucratic attitude.

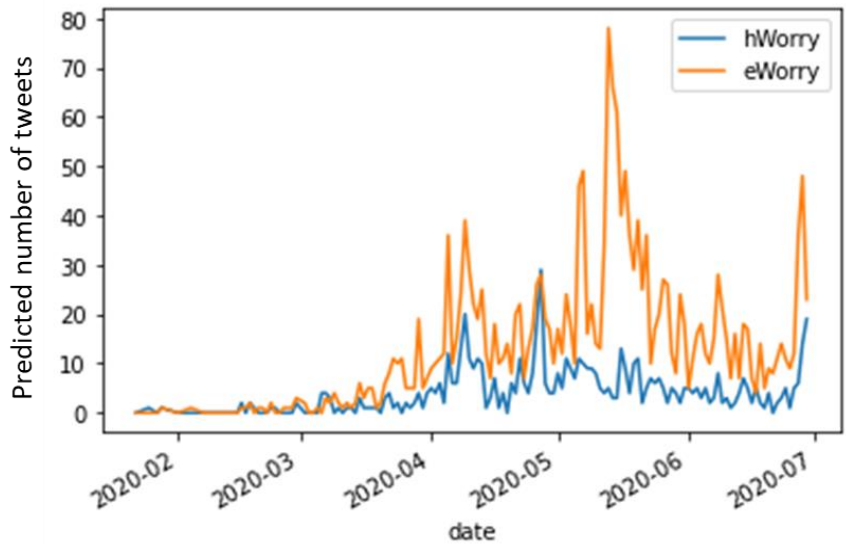

Fig.10. Distribution of the predicted hWorry and eWorry tweets over time

Table 6: Public reactions on the test set using LR with count-vector feature

| No. of tweets | Before LED or ESP | | | After LED or ESP | | |
|---|---|---|---|---|---|---|
| | hWorry | eWorry | Other | hWorry | eWorry | Other |
| Tweets | 87 | 43 | 217 | 165 | 42 | 358 |
| % Tweets | 0.25 | 0.123 | 0.625 | 0.292 | 0.074 | 0.633 |

*Events (LED and ESP) dates for Japan were same

### 5.4. Discussion

We have already discussed the strength and limitations of our method throughout the section 5.

(1) The subseries length, i.e., 15 days before and after each event (LED, ESP) is currently selected by trial and error method. An automated method for such selection will be developed in our future studies.
(2) Although our current results were based on a dataset having approximately 30,000 tweets, we believe in the principle that the more the tweets the better will be the analysis results. Therefore, our current dataset will be augmented with more samples in our future study.
(3) Another point is that the training set we currently used for the classification study are somewhat small and specific for the Japanese tweets. More generalized training set or dictionary and the more sophisticated learning algorithm will be developed in the future studies so that the proposed method can be scaled for more countries.
(4) In addition, we plan to extend our work to the prediction framework which can predict public sentiments during the similar catastrophic situations in future.
(5) The expected realistic comparison of the event effects among multiple countries is troublesome because public reactions in a country depends on its social, economic, political, and cultural conditions, which are usually different in different countries. However, the degree of worries or sensitivity of fundamental issues like economy, health, and politics are well understood among multiple countries considered in this study.

## 6. CONCLUSIONS

We have introduced a method for the even-driven analysis of the public sentiments from corona tweets. This method incorporates a rule-based lexicon for the construction of sentiment timeseries and a machine learning model for the classification of Japanese public sentiments into the "heath-worry" and "economy-worry" classes. To answer how the government responses to the corona pandemic affected the public sentiments, we considered two polarity timeseries (i.e., meanPol and pnRatio) and two events namely LED and ESP. Statistical error analysis on the meanPol and pnRatio sub-series showed a decrease of the negative sentiment after the LED and ESP events for all countries except Japan. Welch's t-test on the above sub-series also showed significant positive impacts on the UK and Australian people by the LED and the USA and UK people by the ESP event, respectively. Manual validation with the relevant tweets approximately showed an agreement with the statistical analysis indicating that the people in different countries have differently affected by the government responses based on their socio-economic and political situations. The proposed logistic regression-based approach classified Japanese public tweets into "heath-worry", "economic-worry" and the "other" classes with 83.11% accuracy (on average). Results showed higher number of tweets on the economy worry, when compared with those with the health worry. An analysis with the classified tweets around each event also re-confirmed the results made by the statistical analysis. Note that the training set (Dataset-2) used for the classification study was extracted independently from the test set (Dataset-1) using health and economy related Japanese keywords. However, more generalized training sets can be developed, which will extend the classification task for multiple countries. Advanced deep learning-based algorithm can also be developed in the future study to perform fine-grained sentiment analysis. A natural extension of our method is to develop algorithm for the prediction of public worry for health and economy during similar disastrous situations in future.

### ACKNOWLEDGEMENTS

We would like to thank all members of the policy data lab of TKFD, who had extended their cordial cooperation during the work in progress.

## AUTHORS

**Md. Khayrul Bashar** received bachelors in electrical and Electronic Engineering from Bangladesh University of Engineering and Technology (BUET), Master of Technology (M. Tech.) in Communication Engineering from Indian Institute of Technology Bombay, and PhD in Information Engineering from Nagoya University in 1993, 1998, and 2004, respectively. After his Ph. D, he served Nagoya University, the University of Tokyo, and Ochanomizu University as a Researcher and faculty member until March 2020. Currently, he is working as a data scientist with Tokyo Foundation for Policy Research. Dr. Bashar published about 20 research articles in the peer-reviewed journals in the fields of image data analysis and algorithms with the application of machine learning algorithms. His research interests include data analytics, applied machine learning, and social computing.